\documentclass[a4paper,fleqn]{cas-dc}
\usepackage[authoryear,round]{natbib}
\bibpunct{(}{)}{;}{a}{,}{,}
\usepackage{cuted}
\usepackage{placeins}

\graphicspath{{./}}
\usepackage{amsmath,amssymb,amsfonts}
\usepackage{booktabs}
\usepackage{multirow}
\usepackage{graphicx}
\usepackage{microtype}
\usepackage{xurl}
\usepackage{bm}
\usepackage{array}
\usepackage{tabularx}
\usepackage{enumitem}
\usepackage{float}
\usepackage{capt-of}
\AtBeginDocument{}

\ExplSyntaxOn
\cs_set:Npn \__make_fig_caption:nn #1#2
{
	\l_fig_align_tl
	\skip_vertical:N \l_fig_abovecap_skip
	
	\setbox\cascaptionbox=\hbox{%
		\sffamily\small
		\textbf{\color{scolor}#1.}~#2
	}
	
	\ifdim\the\wd\cascaptionbox<\dim_use:N \l_fig_width_dim\relax
	\parbox{\l_fig_width_dim}{%
		\unskip\ignorespaces
		\hfil
		\sffamily\small
		\textbf{\color{scolor}#1.}~#2
		\hfil\par
	}
	\else
	\parbox{\l_fig_width_dim}{%
		\rightskip=0pt
		\unskip\ignorespaces
		\sffamily\small
		\textbf{\color{scolor}#1.}~#2\par
	}
	\fi
	
	\skip_vertical:N \l_fig_belowcap_skip
}
\ExplSyntaxOff
\newcommand{\safeincludegraphics}[2][]{%
	\IfFileExists{#2}{%
		\includegraphics[#1]{#2}%
	}{%
		\fbox{%
			\parbox[c][42mm][c]{0.90\linewidth}{%
				\centering\textbf{Figure file not found}\\[0.5em]
				\texttt{\detokenize{#2}}\\[0.5em]
				Place the file in the same folder as the main
				\texttt{.tex} file.%
			}%
		}%
	}%
}

\newcommand{\DIKTtablefont}{\fontsize{9}{11}\selectfont}
\usepackage{xcolor}

\ExplSyntaxOn
\cs_set:Npn \__first_footerline: {}
\ExplSyntaxOff

\definecolor{citationblue}{RGB}{0,120,165}

\AtBeginDocument{%
	\hypersetup{
		colorlinks=true,
		citecolor=citationblue,
		linkcolor=citationblue,
		urlcolor=citationblue
	}
}

\begin{document}
	\let\WriteBookmarks\relax
	
	\setcounter{topnumber}{4}
	\setcounter{bottomnumber}{2}
	\setcounter{totalnumber}{6}
	\setcounter{dbltopnumber}{4}
	\renewcommand{\topfraction}{0.95}
	\renewcommand{\bottomfraction}{0.90}
	\renewcommand{\textfraction}{0.05}
	\renewcommand{\floatpagefraction}{0.80}
	\renewcommand{\dbltopfraction}{0.95}
	\renewcommand{\dblfloatpagefraction}{0.80}
	
	\shorttitle{DIKT: Diagnostic and Interpretable Knowledge Tracing}
	\shortauthors{Wu et~al.}
	\title[mode=title]{DIKT: A Diagnostic and Interpretable Knowledge Tracing Framework for Prediction and Model-based Learning Analysis}
	
	\author[1]{Yi-Qiang Wu}[orcid=0009-0008-4699-3497]
	\fnmark[1]
	\credit{Conceptualization, Methodology, Software, Formal analysis,
    Investigation, Visualization, Writing -- original draft}
	
	\author[2]{Shaohui Chen}[orcid=0009-0004-8707-720X]
	\fnmark[1]
	\credit{Conceptualization, Methodology, Validation, Software,
    Formal analysis, Investigation, Writing -- original draft}
	
	\author[3]{Hanlin Li}[orcid=0000-0002-7908-1127]
	\cormark[1]
	\ead{lihl@wust.edu.cn}
	\credit{Writing -- review and editing}
	
	\author[1]{Baolin Yi}[orcid=0000-0001-8249-9279]
    \cormark[1]
	\ead{epower@mail.ccnu.edu.cn}
	\credit{Writing -- review and editing}
	
	\affiliation[1]{
	organization={Faculty of Artificial Intelligence in Education, Central China Normal University},
	city={Wuhan},
	postcode={430079},
	country={China}
}
	\affiliation[2]{
	organization={School of Mathematics and Systems Science, Wuhan University of Science and Technology},
	city={Wuhan},
	postcode={430065},
	country={China}
}
	\affiliation[3]{
	organization={School of Physics and Mechanics, Wuhan University of Science and Technology},
	city={Wuhan},
	postcode={430065},
	country={China}
}

	\cortext[cor1]{Corresponding author.}
	\fntext[fn1]{The first two authors contributed equally to this work.}

	\begin{abstract}
		Knowledge tracing is a key component of intelligent educational systems. However, most existing methods mainly focus on response prediction, providing limited support for explaining predicted outcomes and improving students' learning states. This paper proposes Diagnostic and Interpretable Knowledge Tracing (DIKT), which integrates response prediction, skill-level diagnosis, and model-based learning analysis. DIKT generates bounded cognitive-state trajectories and decomposes the final response probability into skill-wise probabilities. The trajectories reveal students' evolving strengths and weaknesses, while the skill-wise probabilities identify the skills that most strongly constrain the prediction. Based on these diagnostic results, DIKT supports direct cognitive-state intervention and effective-learning-duration-based what-if analysis to
		examine how hypothetical adjustments affect current predictions and
		subsequent state evolution. DIKT combines Bayesian multi-head attention, conditional variational recurrent inference, a logistic-growth decoder, and multiplicative skill-wise prediction. Experiments on three benchmark datasets show that DIKT outperforms all evaluated baselines, with an average absolute AUC gain of 2.04 percentage points. Other experiments further demonstrate that DIKT supports educationally meaningful diagnostic analysis and provides informative diagnostic indicators for downstream applications, extending knowledge tracing from response prediction to interpretable diagnosis and model-based learning analysis.
	\end{abstract}

	\begin{keywords}
		Knowledge tracing \sep
		Interpretable student modeling \sep
		Skill-level cognitive trajectories \sep
		Skill-level attribution \sep
		Model-based diagnostic analysis
	\end{keywords}
	\begingroup
	\hfuzz=120pt
	\maketitle
	\endgroup
	
	\section{Introduction}
	\label{sec:introduction}
	
	Knowledge tracing (KT) has become an essential component of intelligent educational systems with the rapid development of artificial intelligence and online learning platforms. It aims to infer students' evolving cognitive states from their historical learning interactions and predict their future performance. By providing accurate estimates of students' cognitive states, KT models can support a wide range of educational applications, such as personalized exercise recommendation, learning diagnosis, adaptive tutoring, and remedial instruction.
	
	Over the past decade, substantial research efforts have been devoted to improving the predictive performance of KT models. Early studies mainly relied on probabilistic student models, such as Bayesian Knowledge Tracing, to estimate students' mastery of specific skills. With the development of deep learning, recurrent neural networks, memory-augmented networks, attention mechanisms, and graph-based models have been widely introduced into KT, leading to notable improvements in response prediction. However, most existing deep KT methods still focus primarily on predicting whether a student will correctly answer the next question. Although this prediction-oriented objective is important, it is often insufficient for practical educational decision making.
	
	In real instructional settings, teachers and intelligent tutoring systems require more than an overall prediction score. They also need to determine which skills a student has mastered, which skills are most likely to underlie an incorrect response, and how interventions on the student's current or prior cognitive states may affect subsequent performance. However, many deep knowledge tracing models represent students' cognitive states as implicit latent vectors and produce predictions through largely opaque processes. Consequently, their explanations are often derived passively from internal model mechanisms and are difficult to align with explicit educational semantics. Such explanations therefore provide limited actionable diagnostic information and offer little support for proactively evaluating how hypothetical interventions may alter response outcomes. This substantially restricts the diagnostic value and practical utility of existing knowledge tracing models.
	
	A typical learning scenario further illustrates this limitation. Consider an exercise that primarily assesses two skills, denoted as Skill A and Skill B. Different students may answer the same exercise incorrectly for different reasons: some may fail because of an insufficient cognitive state for Skill A, whereas others may fail because of a weakness in Skill B. Even when students answer the exercise correctly, their underlying cognitive states may still differ. For example, some students may only marginally reach the required cognitive-state levels, whereas others may exhibit substantially higher cognitive states. This phenomenon indicates that a knowledge tracing model limited to response prediction cannot explicitly distinguish among these skill-specific factors, nor can it evaluate how the predicted response would change following a hypothetical improvement in a student's cognitive state for Skill A.

\begin{figure*}[!t]
		\centering
		
		\safeincludegraphics[width=0.78\textwidth]{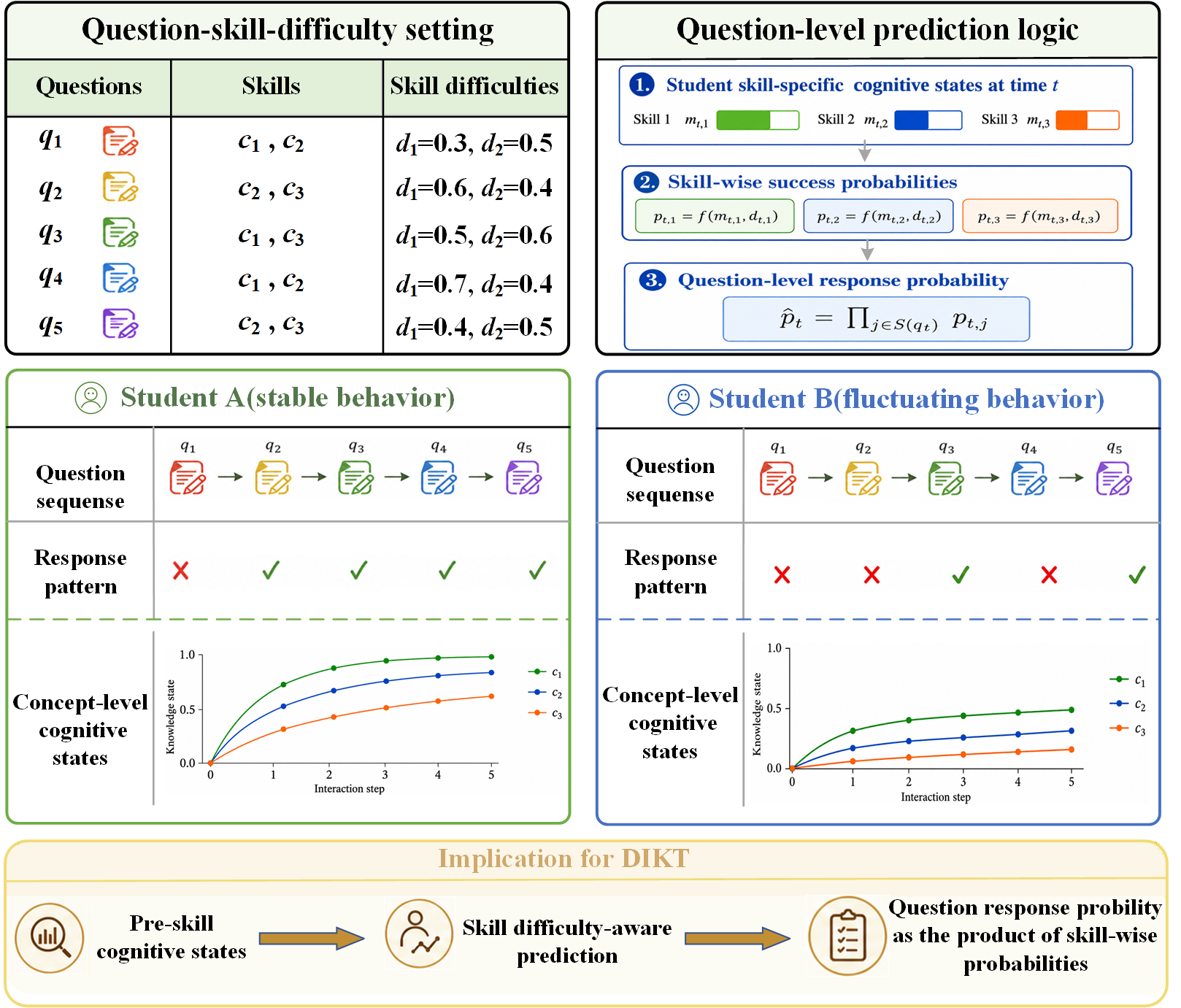}
		
		\vspace{5pt}
		
		\begin{minipage}{0.78\textwidth}
			{\normalfont\small
				
				\refstepcounter{figure}
				\label{fig:research_logic}
				\addcontentsline{lof}{figure}{%
					\protect\numberline{\thefigure}Motivating example for interpretable knowledge tracing.%
				}
				
				\noindent
				\textbf{Fig.~\thefigure.}
				Motivating example for interpretable knowledge tracing. Even under similar exercises, associated skills, and difficulty levels, students may exhibit different response patterns and latent learning dynamics. These differences can lead to distinct skill-level cognitive trajectories, suggesting that response prediction alone is insufficient and motivating explicit interpretable cognitive modeling in knowledge tracing.
				
			}
		\end{minipage}
		
	\end{figure*}

	To address the above limitations, we propose the Diagnostic and
	Interpretable Knowledge Tracing (DIKT) framework, which integrates
	response prediction, skill-level diagnosis, and model-based learning analysis. Fig.~\ref{fig:research_logic} illustrates the motivation and logical pathway of DIKT. An exercise may involve multiple skills, each assessed at a different difficulty level, while a student may exhibit distinct cognitive states for the skills required by that exercise. Accordingly, DIKT estimates a skill-specific success probability for each required skill and obtains the overall probability of answering the exercise correctly by multiplying these skill-wise probabilities. Because student interactions form a temporal sequence, the cognitive state associated with each skill evolves across time steps, thereby forming a skill-level cognitive-state trajectory. These trajectories are used not only to interpret response predictions, but also to support diagnostic and intervention analyses. In this way, DIKT extends knowledge tracing beyond response prediction
	to encompass skill-level diagnosis and model-based learning analysis.

	Specifically, DIKT first employs Bayesian multi-head attention to obtain uncertainty-aware interaction representations. It then uses CVRNN-based latent state inference to capture dynamic learning processes. To generate interpretable cognitive evolution, we further design a logistic-growth cognitive-state decoder and a skill-wise structured prediction layer, which explicitly links skill-level cognitive states to item-level response prediction. Based on the learned cognitive trajectories, DIKT also supports intervention-aware and model-based what-if analysis, allowing the model to explore how hypothetical changes in cognitive states or learning-duration dynamics may affect predictions and future trajectories.
	
	The main contributions of this paper are summarized as follows:
	\begin{itemize}[leftmargin=*]
		\item We propose DIKT, a diagnostic and interpretable knowledge
		tracing framework that jointly supports response prediction,
		skill-level diagnosis, and model-based learning analysis through
		explicit skill-level cognitive states.
		\item We develop an uncertainty-aware sequential inference
		architecture that combines Bayesian multi-head attention with
		CVRNN-based latent-state modeling to capture noisy interaction
		patterns and heterogeneous learning dynamics.
		\item We design a logistic-growth cognitive-state decoder and a
		multiplicative skill-wise prediction mechanism that establish an
		explicit pathway from bounded skill-level cognitive-state estimates
		to skill-wise success probabilities, attribution scores, and the
		final item-level response prediction.
		\item We develop intervention-aware and model-based what-if analysis modules to examine how hypothetical changes in cognitive states or learning-duration dynamics influence predictions and cognitive trajectories.
	\end{itemize}
	
	The remainder of this paper is organized as follows. Section~\ref{sec:relatedwork} reviews related work on deep, interpretable, and diagnostic knowledge tracing. Section~\ref{sec:problem} formulates the knowledge tracing task and defines the explicit skill-level cognitive states considered in this work. Section~\ref{sec:method} introduces the proposed DIKT framework. Section~\ref{sec:experiments} presents the experimental setup and discusses the experimental results. Finally, Section~\ref{sec:conclusion} concludes this paper and outlines future research directions.

	\section{Related Work}
	\label{sec:relatedwork}
	
	This section reviews the studies most closely related to DIKT. We first summarize the development of deep knowledge tracing models and then discuss recent efforts toward interpretable and diagnosis-oriented knowledge tracing.
	
	\subsection{Deep Knowledge Tracing}
	
	Knowledge tracing aims to infer students' evolving knowledge states from historical learning interactions and predict their future performance. Early approaches, such as Bayesian Knowledge Tracing, represented learning through probabilistic transitions and explicit mastery assumptions \citep{corbett1994knowledge}. Several journal surveys have subsequently summarized the evolution of knowledge tracing from probabilistic modeling to deep sequence modeling and have highlighted the growing importance of representation quality, interpretability, cognitive consistency, and educational applicability \citep{abdelrahman2023knowledge,song2022survey,shen2024survey,bai2024survey,yang2025survey}.
	
	Deep Knowledge Tracing introduced recurrent neural networks into KT and demonstrated the effectiveness of neural sequence modeling for response prediction \citep{piech2015deep}. Subsequent studies employed memory networks, self-attention, and contrastive learning to improve student-state representation. DKVMN used an external key--value memory structure to maintain and update latent knowledge states \citep{zhang2017dynamic}, while SKVMN combined sequential modeling with key--value memory to capture longer-term exercise dependencies \citep{abdelrahman2019knowledge}. SAINT and SAKT applied self-attention to model dependencies among learning interactions \citep{choi2020towards,pandey2019self}, and AKT introduced context-aware monotonic attention together with psychometric regularization \citep{ghosh2020context}. KQN further modeled the interaction between a student's knowledge state and the target skill through a knowledge-query mechanism \citep{lee2019knowledge}, whereas CL4KT used pedagogically informed contrastive learning to improve the robustness of learned knowledge representations \citep{lee2022contrastive}. Recent journal-based models have further incorporated exercise information, hierarchical sessions, learning-state diversity, exercise-level concept memory, student--question interaction graphs, and efficient multi-task sequence modeling into student-state estimation \citep{liu2019ekt,ke2024hitskt,wang2025learningstates,mao2025exercise,xu2026sqgkt,liu2026ektml}.
	
	These methods have substantially improved predictive performance. However, most of them represent students' knowledge states using implicit high-dimensional vectors optimized for next-response prediction. Such representations are difficult to associate directly with individual skills and therefore provide limited support for identifying weak skills, tracing interpretable mastery evolution, or conducting targeted instructional analysis.
	
	\subsection{Interpretable and Diagnostic Knowledge Tracing}
	
	Interpretability is essential in educational applications. To enhance
	model transparency, recent KT studies have introduced explicit educational
	structures and explanatory mechanisms
	\citep{abdelrahman2023knowledge,shen2024survey,bai2024survey,yang2025survey}.
	
	Graph- and relation-based methods improve interpretability by explicitly representing relations among students, exercises, and skills. GIKT incorporated high-order question--skill relations into a graph-based interaction model \citep{yang2020gikt}, while heterogeneous-information and heterogeneous-graph models exploited interactions among students, exercises, and skills \citep{xu2023hin,ni2023hhskt}. Recent journal studies have further combined spatiotemporal topological relations, interpretable GCN--GAT structures, multiple knowledge relations, long- and short-term skill graphs, and multi-granularity interaction graphs to capture evolving educational dependencies \citep{li2025sthkt,huo2025gcn,duan2024towards,yu2025exploring,wang2025mgigkt}. Although these models make relational structures more visible, their internal knowledge states are generally not expressed as continuous and directly inspectable skill-level cognitive trajectories.
	
	Question-aware methods provide explanations through cognitive representations, item parameters, or question factors. QIKT used question-centric cognitive representations and an item-response-theory-based prediction mechanism \citep{chen2023improving}; XKT combined cognitive learning theories with multidimensional item response theory \citep{huang2024xkt}; and question-factor-based models incorporated difficulty, discrimination, and response information into prediction \citep{zhao2023exploiting}. Time- and concept-enhanced multidimensional item response theory and cognitive-assimilation modeling further introduced explicit educational parameters and cognitive mechanisms \citep{su2021tcmirt,guo2025icdkt}. Other recent studies modeled knowledge states at both question and skill levels, incorporated metacognitive skills, or aligned implicit predictive states with explicit semantic explanations \citep{li2025interpretable,yin2026metacognitive,ni2026eia}. These methods improve the interpretability of prediction factors and state representations, but they do not necessarily provide an explicit account of how a student's cognitive state of each skill evolves over time and supports controlled intervention analysis.
	
	Learning-process-aware models further incorporate temporal behavior and individual learning characteristics. LPKT directly modeled students' learning processes and learning gains \citep{shen2021learning}, while IEKT considered individual cognition and knowledge-acquisition sensitivity \citep{long2021tracing}. LBKT explicitly modeled the effects of learning behaviors on knowledge-state evolution \citep{xu2023learning}, and DTransformer sought to trace stable knowledge rather than superficial response patterns \citep{yin2023tracing}. Monitoring-based KT modeled continuous learning progress and learning-process consistency \citep{shen2022monitoring}, while TCKT introduced temporal and causal enhancement to obtain more consistent student representations \citep{huang2024learning}. Learning-state-enhanced KT separated knowledge states from broader learning states, and ClusterKT jointly modeled personalized learning gains and forgetting losses through state dependencies \citep{wang2025learningstates,luo2025clusterkt}. More recently, TIAKT introduced a cognitively grounded tri-level memory architecture to model knowledge consolidation, schema formation, and learning transfer \citep{wang2026tiakt}, while quiz-based KT used structured quiz information to model the learning process \citep{shen2025practicing}. These methods improve the consistency and cognitive relevance of learning-process modeling, but their latent states are not necessarily bounded, smooth, or directly interpretable as skill-level cognitive state.
	
	A small number of studies have explored uncertainty, causal relations, and counterfactual-style explanations. UKT represented student interactions using stochastic distribution embeddings and uncertainty-aware attention \citep{cheng2025uncertainty}. IKT incorporated causal relations into interpretable student modeling \citep{minn2022interpretable}, and response influence-based counterfactual reasoning examined how changes in responses affect model outputs \citep{cui2024interpretable}. PSI-KT jointly modeled learner-specific cognitive traits and prerequisite structures through scalable Bayesian inference \citep{zhou2024predictive}, while Markov-blanket-based KT used causal discovery to improve predictive performance and explainability \citep{jiang2024markov}. Nevertheless, these approaches mainly provide uncertainty estimates, interpretable representations, causal-structure information, or post-hoc explanations.
	
	Most existing KT models do not offer an integrated mechanism that converts explicit skill-level cognitive states into actionable diagnosis and targeted intervention analysis. In particular, the connection among cognitive evolution, response prediction, skill diagnosis, and controlled what-if intervention remains insufficiently explored.

	\section{Problem Definition}
	\label{sec:problem}
	
	In a knowledge tracing task, let $\mathcal{S}$, $\mathcal{E}$, and $\mathcal{K}$ denote the sets of students, exercises, and knowledge skills, respectively. For a student, the historical learning sequence is represented as
	\begin{equation}
		X = \{(e_1,y_1),(e_2,y_2),\dots,(e_T,y_T)\},
	\end{equation}
	where $e_t \in \mathcal{E}$ denotes the exercise attempted at time step $t$, 
	and $y_t \in \{0,1\}$ denotes whether the student answers it correctly. 
	Each exercise $e_t$ is associated with one or more knowledge skills, and we denote 
	the corresponding skill set as $\mathcal{K}(e_t) \subseteq \mathcal{K}$.
	
	The standard KT task is to estimate the probability that the student will answer the next exercise correctly:
	\begin{equation}
		\hat{p}_{t+1}
		=
		P(y_{t+1}=1 \mid e_{1:t},y_{1:t},e_{t+1}).
	\end{equation}
	
	To provide skill-level diagnostic information for the predicted
	outcome, DIKT further models an explicit skill-level cognitive state:
	\begin{equation}
		\mathbf{m}_t =
		[m_{t,1},m_{t,2},\dots,m_{t,N_k}]
		\in [0,1]^{N_k},
	\end{equation}
	where $N_k$ is the number of knowledge skills, and $m_{t,k}$
	denotes the model-estimated cognitive state of skill $k$ at time
	step $t$.
	
	For the target exercise $e_{t+1}$, we define the skill-level
	attribution vector as
	\begin{equation}
		\boldsymbol{\rho}^{\mathrm{attr}}_{t+1}
		=
		\left[
		\rho^{\mathrm{attr}}_{t+1,1},
		\rho^{\mathrm{attr}}_{t+1,2},
		\dots,
		\rho^{\mathrm{attr}}_{t+1,J}
		\right]
		\in [0,1]^J,
	\end{equation}
	where $J$ denotes the maximum number of skill slots for one exercise,
	and $\rho^{\mathrm{attr}}_{t+1,j}$ denotes the attribution score
	assigned to the $j$-th required skill slot of the target exercise.
	A larger $\rho^{\mathrm{attr}}_{t+1,j}$ indicates a lower skill-wise
	success probability and a stronger model-estimated constraint on the
	predicted success of the target exercise.
	
	Different from purely prediction-oriented KT methods, DIKT is designed to accurately predict students' future responses while producing interpretable cognitive trajectories and skill-level attribution scores for diagnosis-oriented analysis. The cognitive trajectories characterize the temporal evolution of students' skill-level states, whereas the attribution scores identify the skills that are more likely to constrain the predicted outcome. Based on the learned cognitive states, DIKT further provides model-based intervention-aware and what-if analysis. These analyses are used as probing tools for inspecting model behavior rather than as formal causal identification procedures.
	
	For clarity, the main mathematical notations used throughout this paper are summarized in Table~\ref{tab:notations}.
	\par\medskip
	
	\begin{table}[!t]
		\centering
		
		\begin{minipage}{\columnwidth}
			{\normalfont\DIKTtablefont\raggedright
				
				\refstepcounter{table}
				\label{tab:notations}
				\addcontentsline{lot}{table}{%
					\protect\numberline{\thetable}The key mathematical notations and descriptions.%
				}
				
				\noindent\textbf{Table~\thetable}\par
				\vspace{2pt}
				\noindent The key mathematical notations and descriptions.\par
				\vspace{5pt}
				
				\renewcommand{\arraystretch}{1.02}
				
				\begin{tabularx}{\linewidth}
					{@{}>{\raggedright\arraybackslash}p{0.27\linewidth}
						@{\hspace{4pt}}
						>{\raggedright\arraybackslash}X@{}}
					\toprule
					Notation & Description \\
					\midrule
					
					$e_t$
					& Exercise attempted at time step $t$ \\
					
					$y_t$
					& Binary response label at time step $t$ \\
					
					$\mathcal{K}(e_t)$
					& Skill set associated with exercise $e_t$ \\
					
					$N_k$
					& Number of knowledge skills \\
					
					$J$
					& Maximum number of skill slots for one exercise \\
					
					$\mathbf{x}_t$
					& Interaction representation at time step $t$ \\
					
					$\mathbf{h}_t$
					& Recurrent hidden state at time step $t$ \\
					
					$\mathbf{z}_t$
					& Latent learning variable at time step $t$ \\
					
					$\mathbf{m}_t$
					& Skill-level cognitive-state vector at time step $t$ \\
					
					$m_{t,k}$
					& Estimated cognitive state of skill $k$ at time step $t$ \\
					
					$\alpha_k$
					& Growth-rate parameter of skill $k$ \\
					
					$\beta_k$
					& Offset parameter of skill $k$ \\
					
					$\Delta t_{t,k}$
					& Learning-duration increment of skill $k$ at time step $t$ \\
					
					$\lambda_k$
					& Forgetting rate of skill $k$ \\
					
					$\Delta\tau_t$
					& Idle-duration increment of skill $k$ at time step $t$ \\
					
					$T_{t,k}$
					& Cumulative effective learning duration of skill $k$ \\
					
					$p_{t+1,j}$
					& Skill-wise success probability for the $j$-th required skill slot \\
					
					$\boldsymbol{\rho}^{\mathrm{attr}}_{t+1}$
					& Skill-level attribution vector for target exercise $e_{t+1}$ \\
					
					$\hat{p}_{t+1}$
					& Final predicted response probability for the next exercise \\
					
					$\mathbf{c}_t$
					& Intervention mask for selected cognitive states \\
					
					$\mathbf{m}^{\mathrm{int}}_t$
					& Skill-level cognitive-state in intervention analysis at time step $t$ \\
					
					$\boldsymbol{\delta}^{\mathrm{wf}}$
					& Duration perturbation mask for what-if analysis \\
					
					$\mathbf{m}^{\mathrm{wf}}_t$
					& Skill-level cognitive-state in what-if analysis at time step $t$ \\
					
					\bottomrule
				\end{tabularx}
				
			}
		\end{minipage}
		
	\end{table}
	
	\par\medskip

	\section{The Proposed Method}
	\label{sec:method}
	
	\subsection{Overview of DIKT}
	
\begin{figure*}[!t]
		\centering
		\safeincludegraphics[width=0.78\textwidth]{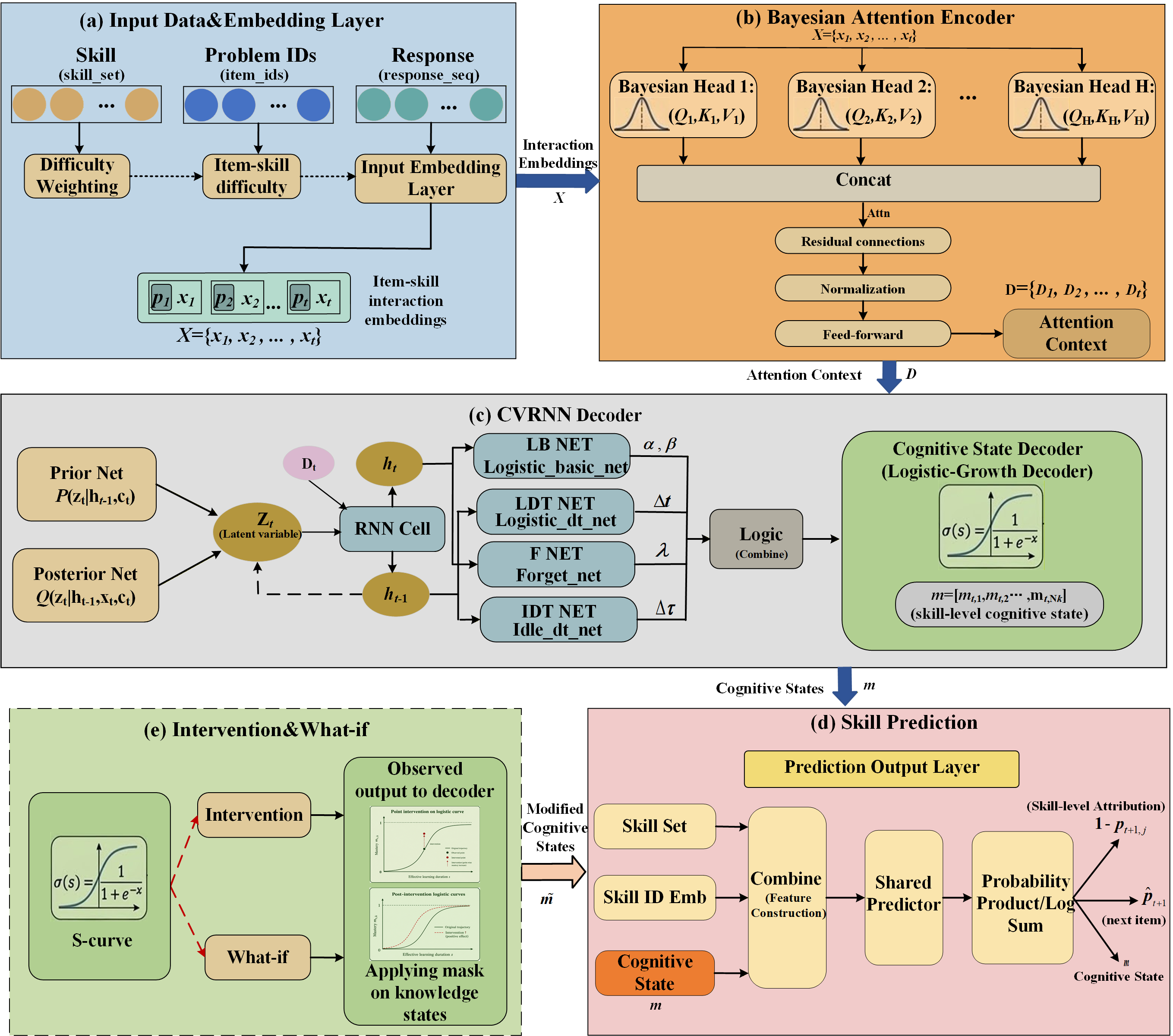}
		\vspace{5pt}
		\begin{minipage}{0.78\textwidth}
			{\normalfont\small
				\centering
				
				\refstepcounter{figure}
				\label{fig:framework1}
				\addcontentsline{lof}{figure}{%
					\protect\numberline{\thefigure}Overall architecture of DIKT.%
				}
				\noindent 
				\textbf{Fig.~\thefigure.}
				Overall architecture of DIKT.
				\par
			}
		\end{minipage}
	\end{figure*}

	To learn interpretable and diagnosis-oriented student representations, we propose Diagnostic and Interpretable Knowledge Tracing (DIKT). The overall architecture is shown in Fig.~\ref{fig:framework1}. As illustrated in the figure, DIKT contains five major components: an input data and embedding layer, a Bayesian attention encoder, a CVRNN-based cognitive-state modeling module, a skill-wise prediction layer, and an intervention and what-if analysis interface. Specifically, the input data and embedding layer integrates exercise, skill, difficulty, and response information into interaction representations. The Bayesian attention encoder captures sequential dependencies and uncertainty-aware interaction patterns. The CVRNN-based module infers dynamic latent learning states and produces explicit skill-level cognitive trajectories, represented by bounded cognitive estimates, through a logistic-growth decoder. The skill-wise prediction layer connects the learned cognitive states with item-level response prediction. The intervention and what-if analysis interface further enables model-based diagnostic analysis by exploring how hypothetical changes in cognitive states may influence predicted responses and subsequent cognitive trajectories.                                  
	
	Compared with conventional deep KT models that mainly rely on implicit hidden representations, DIKT explicitly generates bounded cognitive states over knowledge skills. This design allows the model to be used not only for response prediction, but also for cognitive visualization, diagnostic interpretation, and model-based learning analysis.
	
	\subsection{Interaction Representation and Bayesian Sequence Encoding}
	\label{subsec:encoding}
	
	At each time step, DIKT receives an exercise ID, its associated skills, and the student's response. Since the same response label may correspond to different exercises, skill combinations, and difficulty levels, the input representation should preserve item-specific, skill-specific, and response-related information simultaneously.
	
	For the $j$-th skill associated with exercise $e_t$, the skill ID $s_{t,j}$ is mapped into a dense embedding:
	\begin{equation}
		\mathbf{e}^{(s)}_{t,j}
		=
		\mathrm{Emb}_{skill}(s_{t,j})
		\in \mathbb{R}^{d_h}.
	\end{equation}
	The exercise ID is embedded as
	\begin{equation}
		\mathbf{e}^{(i)}_t
		=
		\mathrm{Emb}_{item}(e_t).
	\end{equation}
	
	To distinguish exercises with different difficulty levels under the same skill, DIKT introduces a learnable difficulty-aware weighting mechanism. For each exercise-skill pair, a difficulty weight is maintained and projected into the hidden space:
	\begin{equation}
		w_{e_t,j} \in [0.1,0.9],
		\qquad
		\mathbf{d}_{t,j}=f_{diff}(w_{e_t,j}).
	\end{equation}
	The skill embedding is then reweighted by
	\begin{equation}
		\tilde{\mathbf{e}}^{(s)}_{t,j}
		=
		\mathbf{e}^{(s)}_{t,j}\odot(1+\mathbf{d}_{t,j}).
	\end{equation}
	
	The skill-based exercise representation is obtained through masked average pooling:
	\begin{equation}
		\mathbf{v}_t=
		\frac{
			\sum_{j=1}^{J}\mathbb{I}(s_{t,j}>0)\tilde{\mathbf{e}}^{(s)}_{t,j}
		}{
			\sum_{j=1}^{J}\mathbb{I}(s_{t,j}>0)+\epsilon
		}.
	\end{equation}
	It is then combined with the exercise embedding:
	\begin{equation}
		\tilde{\mathbf{v}}_t
		=
		f_{qus}([\mathbf{v}_t;\mathbf{e}^{(i)}_t]).
	\end{equation}
	The response label is embedded as
	\begin{equation}
		\mathbf{e}^{(y)}_t
		=
		\mathrm{Emb}_{resp}(y_t),
	\end{equation}
	and the final interaction representation is
	\begin{equation}
		\mathbf{x}_t
		=
		f_{fusion}([\tilde{\mathbf{v}}_t;\mathbf{e}^{(y)}_t]).
	\end{equation}
	
	To preserve temporal order in the interaction sequence, DIKT adds sinusoidal positional encoding to the interaction representation. For the $r$-th hidden dimension, the positional encoding is defined as
	\begin{equation}
		\begin{aligned}
			\mathbf{p}_{t,2r}
			&=\sin\left(\frac{t}{10000^{2r/d_h}}\right),\\
			\mathbf{p}_{t,2r+1}
			&=\cos\left(\frac{t}{10000^{2r/d_h}}\right).
		\end{aligned}
	\end{equation}
	The position-aware interaction representation is then obtained by
	\begin{equation}
		\bar{\mathbf{x}}_t
		=
		\mathbf{x}_t+\mathbf{p}_t.
	\end{equation}
	The resulting position-aware interaction sequence is denoted as
	\begin{equation}
		\bar{\mathbf{X}}
		=
		[\bar{\mathbf{x}}_1,\bar{\mathbf{x}}_2,\ldots,\bar{\mathbf{x}}_T].
	\end{equation}
	
	To capture long-range dependencies and represent uncertainty in student interactions, DIKT uses a Bayesian multi-head attention encoder. Unlike deterministic attention, the projection matrices of query, key, and value are treated as Gaussian-distributed random variables:
	\begin{equation}
		\begin{aligned}
			W_Q &\sim \mathcal{N}(\boldsymbol{\mu}_Q,
			\mathrm{diag}(\boldsymbol{\sigma}_Q^2)),\\
			W_K &\sim \mathcal{N}(\boldsymbol{\mu}_K,
			\mathrm{diag}(\boldsymbol{\sigma}_K^2)),\\
			W_V &\sim \mathcal{N}(\boldsymbol{\mu}_V,
			\mathrm{diag}(\boldsymbol{\sigma}_V^2)).
		\end{aligned}
	\end{equation}
	During training, the projection matrices are sampled using the reparameterization trick:
	\begin{equation}
		W_Q
		=
		\boldsymbol{\mu}_Q
		+
		\boldsymbol{\sigma}_Q\odot\boldsymbol{\epsilon}_Q,
		\qquad
		\boldsymbol{\epsilon}_Q\sim\mathcal{N}(\mathbf{0},\mathbf{I}),
	\end{equation}
	and similarly for $W_K$ and $W_V$. During inference, the posterior means are used.
	
	Given the position-aware interaction sequence $\bar{\mathbf{X}}$, the attention output is computed as
	\begin{equation}
		\begin{aligned}
			\mathrm{Attn}(\bar{\mathbf{X}})
			={}&\,\mathrm{softmax}\left(
			\frac{(\bar{\mathbf{X}}W_Q)
				(\bar{\mathbf{X}}W_K)^\top}
			{\sqrt{d_h/N_h}}
			\right)\\
			&\times(\bar{\mathbf{X}}W_V),
		\end{aligned}
	\end{equation}
	where $N_h$ is the number of attention heads. Residual connections, layer normalization, and feed-forward networks are used to obtain the encoded interaction sequence. During training, the posterior distributions of the Bayesian query, key, and value projection matrices are regularized toward their Gaussian priors through the Bayesian loss defined in Section~\ref{subsec:loss}. This Bayesian formulation reduces the model's over-reliance on a single deterministic attention estimate and improves the robustness of sequence encoding for sparse and noisy educational interaction data.

	\subsection{Variational Skill-Level Cognitive-State Modeling}
	\label{subsec:mastery}
	
	The encoded interaction representations are then fed into a conditional variational recurrent module to infer dynamic latent learning states. This module is introduced because students with similar observed responses may still differ in hidden learning pace, retention tendency, or short-term learning instability.
	
	At time step $t$, the latent variable $\mathbf{z}_t$ is modeled with a prior distribution:
	\begin{equation}
		p(\mathbf{z}_t|\mathbf{h}_{t-1})
		=
		\mathcal{N}
		(\boldsymbol{\mu}^{prior}_t,
		\mathrm{diag}((\boldsymbol{\sigma}^{prior}_t)^2)),
	\end{equation}
	and a posterior distribution:
	\begin{equation}
		q(\mathbf{z}_t|\mathbf{x}_t,\mathbf{h}_{t-1})
		=
		\mathcal{N}
		(\boldsymbol{\mu}^{post}_t,
		\mathrm{diag}((\boldsymbol{\sigma}^{post}_t)^2)).
	\end{equation}
	During training, $\mathbf{z}_t$ is sampled from the posterior:
	\begin{equation}
		\mathbf{z}_t
		=
		\boldsymbol{\mu}^{post}_t
		+
		\boldsymbol{\sigma}^{post}_t
		\odot
		\boldsymbol{\epsilon},
		\qquad
		\boldsymbol{\epsilon}\sim\mathcal{N}(0,I).
	\end{equation}
	The recurrent hidden state is updated by
	\begin{equation}
		\mathbf{h}_t
		=
		\mathrm{GRUCell}([\mathbf{x}_t;\mathbf{z}_t],\mathbf{h}_{t-1}).
	\end{equation}
	
	Instead of directly mapping $\mathbf{h}_t$ to an unconstrained cognitive vector, DIKT generates skill-specific growth parameters:
	\begin{equation}
		[\alpha_{1},\dots,\alpha_{N_k},
		\beta_{1},\dots,\beta_{N_k}]
		=
		f_{basic}(\mathbf{h}_T),
	\end{equation}
	where $\alpha_{k}$ and $\beta_{k}$ are bounded to $(0,1)$ through a sigmoid layer. The model also estimates an effective learning-duration increment:
	\begin{equation}
		\Delta t_{t,k}
		=
		f_{\Delta t}([\mathbf{z}_t;\mathbf{h}_{t-1}])_k.
	\end{equation}
	
	The cognitive state is generated through a recursive logistic-growth
	decoding process. Let $T_{t-1,k}$ denote the equivalent cumulative
	effective learning duration of skill $k$ after the previous update.
	At time step $t$, the effective learning duration is first updated as
	\begin{equation}
		\widetilde{T}_{t,k}
		=
		T_{t-1,k}
		+
		\Delta t_{t,k},
	\end{equation}
	where $T_{0,k}=0$. The corresponding preliminary cognitive state is
	then generated by the logistic-growth function:
	\begin{equation}
		\widetilde{m}_{t,k}
		=
		\frac{1}
		{1+\exp\left(
			-\alpha_{k}
			(\widetilde{T}_{t,k}-\beta_{k})
			\cdot\gamma
			\right)},
	\end{equation}
	where $\gamma$ is a scaling constant.
	
	After the learning update, the preliminary cognitive state is adjusted
	through an exponential decay function:
	\begin{equation}
		m_{t,k}
		=
		\widetilde{m}_{t,k}
		\exp\left(
		-\lambda_k\Delta\tau_t
		\right).
	\end{equation}
	
	The skill-specific forgetting rate is generated as
	\begin{equation}
		\lambda_k
		=
		\left[
		f_{\mathrm{forget}}(\mathbf{h}_T)
		\right]_k
		\in(0,1),
	\end{equation}
	where $\lambda_k$ denotes the learned forgetting rate of skill $k$.
	The model-estimated idle-duration increment at time step $t$ is defined as
	\begin{equation}
		\Delta\tau_t
		=
		f_{\mathrm{idle}}
		\left(
		[\mathbf{z}_t;\mathbf{h}_{t-1}]
		\right).
	\end{equation}
	
	The exponential decay term simulates forgetting by reducing the
	preliminary cognitive state according to the skill-specific forgetting
	rate and the estimated idle duration. 
	
	The corrected cognitive state is subsequently converted back into an
	equivalent cumulative effective learning duration through the inverse
	logistic transformation:
	\begin{equation}
		T_{t,k}
		=
		\beta_{k}
		-
		\frac{
			\log\left(
			\frac{1}{m_{t,k}}-1+\epsilon
			\right)
		}{
			\alpha_{k}\gamma
		}.
	\end{equation}
	The resulting $T_{t,k}$ is carried forward as the initial effective
	learning duration for the next time step. This recursive process
	constrains the cognitive states to $(0,1)$ and preserves the smooth and
	saturating pattern of the logistic-growth trajectory, while allowing
	the estimated states to exhibit gradual decreases during idle periods.
	
	\subsection{Skill-Wise Prediction and Model-Based Analysis}
	\label{subsec:prediction_analysis}
	
	After obtaining the cognitive state $\mathbf{m}_t$, DIKT predicts the student's response to the next exercise. For the $j$-th skill slot of the target exercise $e_{t+1}$, the model constructs a skill-wise prediction feature:
	\begin{equation}
		\mathbf{f}_{t+1,j}
		=
		[\tilde{\mathbf{v}}_{t+1};\,m_{t,s_{t+1,j}};\,\mathbf{u}_{t+1,j}],
	\end{equation}
	where $\mathbf{u}_{t+1,j}$ is the skill-ID embedding. A shared skill predictor produces the corresponding skill-wise success probability:
	\begin{equation}
		p_{t+1,j}
		=
		\sigma(f_{skill}(\mathbf{f}_{t+1,j})).
	\end{equation}
	
	The corresponding skill-level attribution score is computed as
	\begin{equation}
		\rho^{\mathrm{attr}}_{t+1,j}
		=
		1-p_{t+1,j},
		\label{eq:skill_attribution}
	\end{equation}
	
	The final response probability is obtained through multiplicative fusion:
	\begin{equation}
		\hat{p}_{t+1}
		=
		\prod_{j:s_{t+1,j}>0}p_{t+1,j}.
	\end{equation}
	For numerical stability, it is implemented as
	\begin{equation}
		\begin{aligned}
			\log \hat{p}_{t+1}
			&=
			\sum_{j:s_{t+1,j}>0}
			\log(p_{t+1,j}+\epsilon),\\
			\hat{p}_{t+1}
			&=
			\exp(\log \hat{p}_{t+1}).
		\end{aligned}
	\end{equation}
	
	The multiplicative fusion is used as a structured aggregation mechanism
	rather than as a strict causal equation. Its purpose is to preserve an
	explicit pathway from skill-level cognitive-state estimates to item-level
	response prediction. The skill-wise success probabilities provide the
	computational basis for the attribution scores, while their multiplicative
	aggregation produces the final item-level response probability.
	
	DIKT further supports mask-based attribution exclusion. Let
	$\mathbf{R}=[R_1,\ldots,R_J]\in\{0,1\}^{J}$ denote the exclusion mask,
	where $R_j=0$ retains the $j$-th skill and $R_j=1$ excludes it. The
	masked skill-wise probability is defined as
	\begin{equation}
		\widetilde{p}_{t+1,j}
		=
		(1-R_j)p_{t+1,j}+R_j,
	\end{equation}
	so that $\widetilde{p}_{t+1,j}=1$ when $R_j=1$.
	
	Based on the learned cognitive states, DIKT further provides two model-based analysis modules.
	
	First, the intervention-aware module adjusts selected cognitive dimensions through a mask-based intervention operation. Let
	\begin{equation}
		\mathbf{c}_t
		=
		[c_{t,1},c_{t,2},\dots,c_{t,N_k}]
	\end{equation}
	denote the intervention mask at time step $t$, where each entry satisfies
	\begin{equation}
		c_{t,k}\in \{-1\}\cup[0,1].
	\end{equation}
	The default value $c_{t,k}=-1$ indicates that no intervention is applied to skill $k$. If $c_{t,k}\ge 0$, the cognitive value of skill $k$ is directly replaced by the specified intervention value. Formally, the intervened cognitive state is defined as
	\begin{equation}
		m^{\mathrm{int}}_{t,k}
		=
		\begin{cases}
			m_{t,k}, & c_{t,k}<0,\\
			c_{t,k}, & c_{t,k}\ge 0.
		\end{cases}
	\end{equation}
	This masking strategy allows DIKT to intervene only on selected skills while keeping all other skill-level cognitive states unchanged. The intervened cognitive state $\mathbf{m}^{\mathrm{int}}_t$ is then passed to the same skill-wise prediction layer. By varying the non-negative entries in $\mathbf{c}_t$ for selected skills, DIKT can produce intervention-response curves that show how hypothetical cognitive changes affect the predicted response probability under the learned model.
	
	Second, the what-if module perturbs the effective learning-duration dynamics rather than directly overwriting cognitive values. Let $L$ denote the length of the learning sequence. We define the temporal-skill mask as
	\begin{equation}
		\boldsymbol{\delta}^{\mathrm{wf}}
		=
		[\delta^{\mathrm{wf}}_{\tau,k}]
		\in
		\mathbb{R}^{L\times N_k},
	\end{equation}
	where $\delta^{\mathrm{wf}}_{\tau,k}$ represents the perturbation applied to the effective duration increment of skill $k$ at time step $\tau$. To ensure that the perturbed duration remains valid, the perturbation is constrained by
	\begin{equation}
		\Delta t_{\tau,k}
		+
		\delta^{\mathrm{wf}}_{\tau,k}
		\ge 0.
	\end{equation}
	The perturbed duration increment is then computed as
	\begin{equation}
		\Delta t^{\mathrm{wf}}_{\tau,k}
		=
		\Delta t_{\tau,k}
		+
		\delta^{\mathrm{wf}}_{\tau,k}.
	\end{equation}
	
	The what-if cognitive state $m^{\mathrm{wf}}_{t,k}$ is then generated using the same logistic-growth decoding process defined above, with
	$\Delta t_{\tau,k}$ replaced by $\Delta t^{\mathrm{wf}}_{\tau,k}$ at each time step. The resulting sequence of cognitive states constitutes the corresponding what-if cognitive trajectory.
	
	Both modules are intended as model-based probing tools for instructional analysis. The mask-based formulation makes the hypothetical operation explicit and prevents unrelated skills or time steps from being unintentionally modified. Therefore, the intervention-aware module analyzes the immediate effect of directly modifying selected cognitive values, whereas the what-if module examines how changes in learning-duration dynamics may influence the subsequent cognitive trajectory.
	
	\subsection{Training Objective}
	\label{subsec:loss}
	
	DIKT is trained end-to-end using a multi-term objective that jointly optimizes response prediction, Bayesian parameter regularization, latent-state inference, and the interpretability-related properties of the cognitive trajectories:
	\begin{equation}
		\begin{aligned}
			\mathcal{L}
			=&\,
			\mathcal{L}_{pred}
			+
			\lambda_{attn}\mathcal{L}_{attn}
			+
			\lambda_{Bayes}\mathcal{L}_{Bayes} \\
			&+
			\lambda_{VAE}\mathcal{L}_{VAE}
			+
			\lambda_{mono}\mathcal{L}_{mono}
			+
			\lambda_{scale}\mathcal{L}_{scale},
		\end{aligned}
	\end{equation}
	where $\lambda_{attn}$, $\lambda_{Bayes}$, $\lambda_{VAE}$,
	$\lambda_{mono}$, and $\lambda_{scale}$ control the contributions of
	the corresponding auxiliary and regularization terms.
	
	The main prediction loss is the binary cross-entropy between the
	predicted and observed next responses:
	\begin{equation}
		\mathcal{L}_{pred}
		=
		\sum_{t=1}^{T-1}
		\mathrm{BCE}(y_{t+1},\hat{p}_{t+1}).
	\end{equation}
	
	An auxiliary attention-based prediction head is used to stabilize
	the learning of sequential interaction representations:
	\begin{equation}
		\mathcal{L}_{attn}
		=
		\sum_{t=1}^{T-1}
		\mathrm{BCE}(y_{t+1},\hat{p}^{attn}_{t+1}).
	\end{equation}
	
	The Bayesian attention loss regularizes the posterior distributions of the query, key, and value projection matrices toward their Gaussian priors:
	
	\begin{equation}
		\begin{aligned}
			\mathcal{L}_{\mathrm{Bayes}}
			={}&
			\sum_{\theta\in\{Q,K,V\}}
			\mathrm{KL}\Bigl(
			\mathcal{N}\bigl(
			\boldsymbol{\mu}_{\theta},
			\mathrm{diag}(\boldsymbol{\sigma}_{\theta}^{2})
			\bigr)\\
			&\qquad\Big\|
			\mathcal{N}\bigl(
			\mathbf{0},
			\sigma_{\mathrm{prior}}^{2}\mathbf{I}
			\bigr)\Bigr).
		\end{aligned}
	\end{equation}

	This term constrains the posterior distributions of the Bayesian attention parameters to remain close to the predefined Gaussian prior, thereby reducing overfitting and improving the robustness of attention estimation under sparse and noisy student interaction data.

	The variational latent-state regularization is defined as
	\begin{equation}
		\mathcal{L}_{VAE}
		=
		\frac{1}{T}
		\sum_{t=1}^{T}
		\mathrm{KL}
		\left(
		q(\mathbf{z}_t|\mathbf{x}_t,\mathbf{h}_{t-1})
		\;\middle\|\;
		p(\mathbf{z}_t|\mathbf{h}_{t-1})
		\right).
	\end{equation}

	To ensure that the predicted probability is positively consistent with the cognitive state of the corresponding skill, DIKT introduces a monotonicity regularization term for the skills associated with the target exercise. The regularization consists of three parts: a first-order monotonicity constraint, a lower-boundary constraint, and an upper-boundary constraint.
	
	For each skill $k \in \mathcal{K}(e_{t+1})$, let $p_{t+1,k}(m_{t,k})$ denote the skill-wise prediction probability produced by the shared skill predictor when the cognitive state of skill $k$ is set to $m_{t,k}$, while other inputs remain unchanged. The first-order monotonicity constraint is defined as
	\begin{equation}
		\mathcal{L}_{\mathrm{grad}}
		=
		\mathbb{E}
		\left[
		\frac{1}{|\mathcal{K}(e_{t+1})|}
		\sum_{k\in \mathcal{K}(e_{t+1})}
		\mathrm{ReLU}
		\left(
		-\frac{\partial p_{t+1,k}}
		{\partial m_{t,k}}
		+\epsilon
		\right)
		\right].
	\end{equation}
	
	To further encourage meaningful boundary behavior, DIKT constrains the skill-wise prediction probability to approach 0 when the corresponding cognitive state is 0, and to approach 1 when the corresponding cognitive state is 1:
	\begin{equation}
		\mathcal{L}_{\mathrm{low}}
		=
		\mathbb{E}
		\left[
		\frac{1}{|\mathcal{K}(e_{t+1})|}
		\sum_{k\in \mathcal{K}(e_{t+1})}
		\left(
		p_{t+1,k}(0)
		\right)^2
		\right],
	\end{equation}
	\begin{equation}
		\mathcal{L}_{\mathrm{high}}
		=
		\mathbb{E}
		\left[
		\frac{1}{|\mathcal{K}(e_{t+1})|}
		\sum_{k\in \mathcal{K}(e_{t+1})}
		\left(
		1-
		p_{t+1,k}(1)
		\right)^2
		\right].
	\end{equation}
	
	The complete monotonicity regularization term is then defined as
	\begin{equation}
		\mathcal{L}_{\mathrm{mono}}
		=
		\mathcal{L}_{\mathrm{grad}}
		+
		\eta_{\mathrm{low}}
		\mathcal{L}_{\mathrm{low}}
		+
		\eta_{\mathrm{high}}
		\mathcal{L}_{\mathrm{high}},
	\end{equation}
	where $\eta_{\mathrm{low}}$ and $\eta_{\mathrm{high}}$ control the strengths of the two boundary constraints.
	
	This regularization encourages the skill-wise prediction probability to increase monotonically with the corresponding cognitive state, while also promoting meaningful boundary behavior at the two extremes of the cognitive-state range. As a result, the prediction layer is encouraged to follow an educationally consistent mapping in which a low cognitive state corresponds to a low success probability, whereas a high cognitive state corresponds to a high success probability.

	Finally, DIKT introduces scale and smoothness regularization to constrain the range and temporal evolution of the cognitive trajectories:
	\begin{equation}
		\begin{aligned}
			\mathcal{L}_{\mathrm{scale}}
			=&\,
			(1-\max(\mathbf{m}))^2
			+
			\frac{1}{N_k}
			\sum_{k=1}^{N_k}(m_{1,k})^2 \\
			&+
			\frac{1}{T-1}
			\sum_{t=1}^{T-1}
			\left\|
			\mathbf{m}_{t+1}-\mathbf{m}_t
			\right\|_2^2 \\
			&+
			\frac{1}{T-2}
			\sum_{t=1}^{T-2}
			\left\|
			(\mathbf{m}_{t+2}-\mathbf{m}_{t+1})
			-
			(\mathbf{m}_{t+1}-\mathbf{m}_t)
			\right\|_2^2 .
		\end{aligned}
	\end{equation}
	
	The first two terms constrain the overall scale and initial values of the cognitive states. The third term penalizes abrupt first-order changes between adjacent time steps, while the fourth term penalizes rapid changes in the trajectory growth rate. Together, these terms encourage bounded, smooth, and educationally plausible cognitive trajectories.

	\section{Experiments}
	\label{sec:experiments}
	
	In this section, we conduct experiments to evaluate DIKT from both predictive and diagnostic perspectives. We first introduce the experimental setup, including datasets, baselines, evaluation metrics, and implementation details. Then, we report the experimental results and discuss the effectiveness of DIKT by answering the following research questions:
	
	\begin{itemize}[leftmargin=*]
		\item \textbf{RQ1:} Can DIKT achieve competitive predictive performance compared with representative KT baselines?
		\item \textbf{RQ2:}  Can DIKT generate interpretable skill-level cognitive trajectories and skill-level attribution scores?
		\item \textbf{RQ3:} Can the learned cognitive states support model-based intervention-aware and what-if analysis?
		\item \textbf{RQ4:} How do the main components of DIKT contribute to prediction and interpretability?
	\end{itemize}
	
	\subsection{Experimental Setup}
	\label{subsec:experimental_setup}
	
	\subsubsection{Datasets}
	\label{subsubsec:datasets}
	
	We evaluate DIKT on three public benchmark datasets for knowledge tracing: ASSIST09, ASSIST17, and Algebra2005. ASSIST09 and ASSIST17 are collected from the ASSISTments platform, whereas Algebra2005 is collected from algebra learning records in an intelligent tutoring environment. These datasets vary in terms of the numbers of students, exercises, skills, and interactions, which enables us to examine the robustness of DIKT under different educational scenarios and sequence characteristics.

	The statistics in Table~\ref{tab:dataset_stats} are reported on the preprocessed versions used in this paper. Following common KT preprocessing practice, interaction records are organized into student-level sequences, and exercises without valid skill annotations are removed.
	
	\par\medskip
	
	\begin{table*}[!t]
		\centering
		\begin{minipage}{0.96\textwidth}
			{\normalfont\small\raggedright
				
				\refstepcounter{table}
				\label{tab:dataset_stats}
				\addcontentsline{lot}{table}{%
					\protect\numberline{\thetable}Statistics of the datasets used in the experiments.%
				}
				
				\noindent\textbf{Table~\thetable}\par
				\vspace{2pt}
				\noindent Statistics of the datasets used in the experiments.\par
				\vspace{5pt}
				
				\renewcommand{\arraystretch}{1.05}
				\begin{tabular*}{\linewidth}{@{\extracolsep{\fill}}lccccc@{}}
					\toprule
					Dataset & Students & Exercises & Skills & Interactions & Avg. Length \\
					\midrule
					ASSIST09    & 4,217 & 26,688  & 123 & 346,860 & 82.25 \\
					ASSIST17    & 1,709 & 3,162   & 102 & 942,816 & 551.68 \\
					Algebra2005 & 574   & 210,710 & 112 & 809,694 & 1410.62 \\
					\bottomrule
				\end{tabular*}
				
			}
		\end{minipage}
	\end{table*}
	
	\par\medskip

	\subsubsection{Baselines}
	\label{subsubsec:baselines}
	
	We compare DIKT with representative KT baselines from different modeling families:
	
	\begin{itemize}[leftmargin=*]
		\item \textbf{BKT} \citep{corbett1994knowledge}: a classical probabilistic KT model based on hidden Markov assumptions;
		\item \textbf{DKT} \citep{piech2015deep}: a recurrent neural network based KT model for student response sequence modeling;
		\item \textbf{SAINT} \citep{choi2020towards}: a Transformer-based KT model that applies self-attention to exercise and response sequences;
		\item \textbf{QIKT} \citep{chen2023improving}: an interpretable KT model that introduces question-centric cognitive representations;
		\item \textbf{DSN-KT} \citep{gao2025dsnkt}: a recent
		deep semantics-robust knowledge tracing model that combines knowledge augmentation, time-cumulative attention, and semantic contrastive learning to obtain stable knowledge-state representations;
		\item \textbf{KQN} \citep{lee2019knowledge}: a knowledge query network model that predicts student performance through skill-specific knowledge representations;
		\item \textbf{SAKT} \citep{pandey2019self}: a self-attentive KT model that captures relevant historical interactions for response prediction.
	\end{itemize}
	
	These baselines cover probabilistic KT, recurrent neural KT, Transformer-based KT, interpretable KT, knowledge-query-based KT, and self-attention-based KT. This comparison allows us to examine whether DIKT can maintain competitive predictive performance while providing explicit skill-level cognitive interpretation and model-based diagnostic analysis.

	\subsubsection{Metrics}
	\label{subsubsec:metrics}
	
	For predictive performance, we use AUC and ACC as the main evaluation metrics, both of which are commonly adopted for evaluating predictive performance in knowledge tracing. AUC measures the model's ability to rank positive responses higher than negative responses across different classification thresholds, while ACC measures the proportion of correctly predicted student responses.

	In addition to prediction metrics, DIKT also supports diagnostic analysis,
	including intervention and what-if analysis. The intervention and what-if metrics are not used as primary ranking criteria. Instead, they are used to inspect whether the learned cognitive states can support model-based diagnostic analysis.
	
	For intervention analysis, we evaluate how the predicted response
	probability changes when the cognitive value of a selected skill is
	hypothetically adjusted. For a target skill $k$ at time step $t$, the
	original cognitive value $m_{t,k}$ is replaced by a specified intervention
	value $v\in[0,1]$, while the cognitive values of all unselected skills
	remain unchanged. Let $\hat{p}_{t+1}$ denote the original predicted
	response probability and let $\hat{p}^{\mathrm{int}}_{t+1}(v)$ denote
	the prediction obtained after the single-point intervention. The
	point-wise intervention efficiency is defined as
	\begin{equation}
		\mathrm{Eff}_{t,k}(v)
		=
		\frac{
			\left|
			\hat{p}^{\mathrm{int}}_{t+1}(v)
			-
			\hat{p}_{t+1}
			\right|
		}{
			\left|
			v-m_{t,k}
			\right|
			+\epsilon
		}.
	\end{equation}
	Here, $\epsilon$ is a small positive constant used for numerical
	stability. A larger value of $\mathrm{Eff}_{t,k}(v)$ indicates that
	the predicted response probability is more sensitive to a unit change
	in the selected skill cognitive value. 
	
	For what-if analysis, we quantify the difference between the original and perturbed cognitive trajectories using the normalized $L_2$ distance:
	\begin{equation}
		D_{L2}
		=
		\sqrt{
			\frac{1}{T N_k}
			\sum_{t=1}^{T}
			\sum_{k=1}^{N_k}
			\left(
			m^{\mathrm{wf}}_{t,k}-m_{t,k}
			\right)^2
		}.
	\end{equation}
	A larger $D_{L2}$ indicates a greater overall change in the skill-level cognitive trajectory after the hypothetical perturbation.

	\subsubsection{Implementation}
	\label{subsubsec:implementation}
	
	DIKT is implemented in PyTorch and trained using the Adam optimizer. The learning rate is set to 0.001 and the batch size is set to 16. The main hyperparameters are summarized in Table~\ref{tab:hyperparams}.
	
	\begin{table}[!htbp]
		\centering
		
		\begin{minipage}{0.95\linewidth}
			{\normalfont\small
				
				\refstepcounter{table}
				\label{tab:hyperparams}
				\addcontentsline{lot}{table}{%
					\protect\numberline{\thetable}Implementation details and hyperparameter settings.%
				}
				
				\noindent\textbf{Table~\thetable}\par
				\vspace{2pt}
				\noindent Implementation details and hyperparameter settings.\par
				\vspace{5pt}
				
				\renewcommand{\arraystretch}{1.08}
				
				\begin{tabular*}{\linewidth}{@{\extracolsep{\fill}}lr@{}}
					\toprule
					Hyperparameter & Value \\
					\midrule
					Hidden dimension & 64 \\
					Number of attention heads & 2 \\
					Number of attention layers & 2 \\
					Latent dimension & 16 \\
					Dropout rate & 0.1 \\
					Learning rate & 0.001 \\
					Batch size & 16 \\
					$\lambda_{\mathrm{Bayes}}$ & 0.1 \\
					$\lambda_{\mathrm{VAE}}$ & 0.1 \\
					$\lambda_{\mathrm{attn}}$ & 0.5 \\
					$\lambda_{\mathrm{mono}}$ & 0.5 \\
					$\lambda_{\mathrm{scale}}$ & 1.0 \\
					\bottomrule
				\end{tabular*}
				
			}
		\end{minipage}
	\end{table}

	\begin{table*}[!t]
		\centering
		\begin{minipage}{0.92\textwidth}
			{\normalfont\small\raggedright
				
				\refstepcounter{table}
				\label{tab:main_results}
				\addcontentsline{lot}{table}{%
					\protect\numberline{\thetable}Predictive performance comparison on benchmark datasets.%
				}
				
				\noindent\textbf{Table~\thetable}\par
				\vspace{2pt}
				\noindent Predictive performance comparison on benchmark datasets.\par
				\vspace{5pt}
				
				\renewcommand{\arraystretch}{1.08}
				
				\begin{tabular*}{\linewidth}{@{\extracolsep{\fill}}lcccccc@{}}
					\toprule
					\multirow{2}{*}{Methods}
					& \multicolumn{2}{c}{ASSIST09}
					& \multicolumn{2}{c}{ASSIST17}
					& \multicolumn{2}{c}{Algebra2005} \\
					\cmidrule(lr){2-3}
					\cmidrule(lr){4-5}
					\cmidrule(lr){6-7}
					& AUC & ACC & AUC & ACC & AUC & ACC \\
					\midrule
					BKT   & 0.6511 & 0.6342 & 0.6017 & 0.5874 & 0.6500 & 0.7553 \\
					DKT   & 0.7561 & 0.7362 & 0.7178 & 0.7098 & 0.8149 & 0.8097 \\
					SAINT & 0.7441 & 0.7358 & 0.7087 & 0.6922 & 0.7775 & 0.7791 \\
					QIKT  & 0.7801 & 0.7377 & 0.7645 & 0.6985 & 0.8408 & 0.8222 \\
					DSN-KT & 0.8016 & 0.7546 & 0.6886 & 0.6811 & 0.7793 & 0.8354 \\
					KQN   & 0.7546 & 0.7249 & 0.7065 & 0.6611 & 0.8027 & 0.8025 \\
					SAKT  & 0.7861 & 0.7681 & 0.7261 & 0.7144 & 0.7880 & 0.7954 \\
					\midrule
					DIKT
					& \textbf{0.8272}
					& \textbf{0.7761}
					& \textbf{0.7925}
					& \textbf{0.7313}
					& \textbf{0.8485}
					& \textbf{0.8598} \\
					\bottomrule
				\end{tabular*}
				
			}
		\end{minipage}
	\end{table*}
	
	\subsection{Performance Comparison (RQ1)}
	\label{subsec:performance_comparison}
	
	This subsection compares DIKT with representative KT baselines and
	examines whether explicit skill-level cognitive-state modeling can be
	introduced without compromising response prediction performance.
	
	As shown in Table~\ref{tab:main_results}, DIKT achieves competitive predictive performance on the benchmark datasets. On ASSIST09, DIKT obtains an AUC of 0.8272, outperforming the representative baselines listed in the table. This result indicates that explicit skill-level cognitive modeling does not necessarily weaken prediction performance. Instead, the learned cognitive states can provide useful information for response prediction.
	
	The comparison also suggests that DIKT should not be evaluated only as a prediction-oriented KT model. Unlike methods that mainly optimize the next-response objective, DIKT is designed to jointly support prediction, cognitive state interpretation, and model-based learning analysis. Therefore, the following experiments further examine whether the learned states have educationally meaningful properties.
	
	\subsection{Cognitive State and Attribution Analysis (RQ2)}
	\label{subsec:mastery_analysis}
	
	This subsection evaluates the interpretability of DIKT from two
	complementary perspectives: the temporal evolution of skill-level
	cognitive states and the exercise-specific attribution of the skills
	required by each exercise.
	
	We first examine whether the skill-level cognitive states generated by
	DIKT are bounded, smooth, and distinguishable across skills. It should
	be noted that these cognitive states are model-inferred probabilistic
	representations rather than directly observable variables. Therefore,
	the generated trajectories are used to characterize and visualize the
	estimated evolution of students' skill-level knowledge under the
	modeling assumptions of DIKT.
	
	\par
	\vspace{5pt}
	\noindent
	\begin{minipage}{\columnwidth}
		\centering
		\safeincludegraphics[
		width=0.89\linewidth
		]{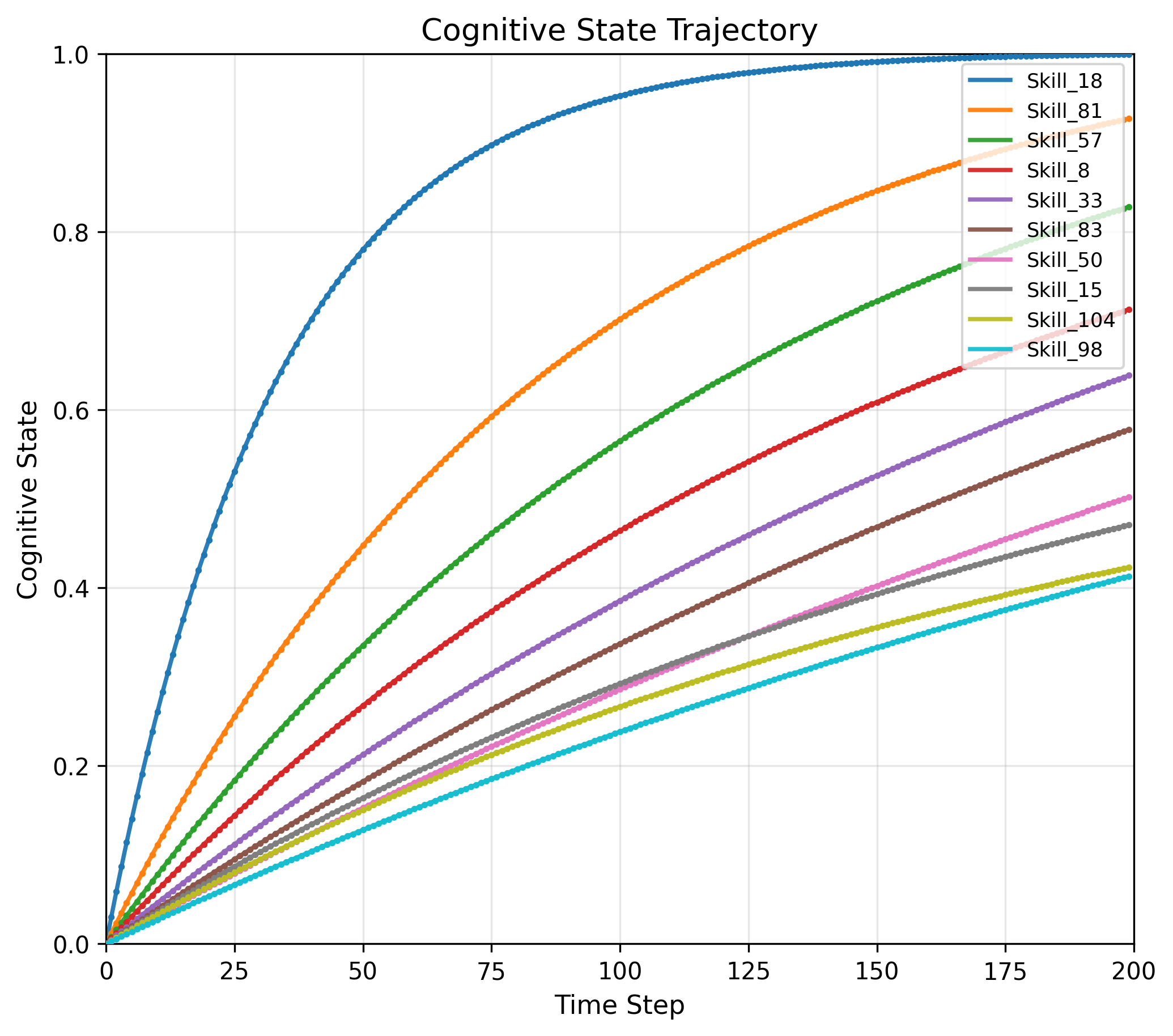}
		\vspace{2pt}
		
		{\normalfont\small
			\setlength{\parindent}{0pt}
			\refstepcounter{figure}
			\label{fig:knowledge_trajectory}
			\addcontentsline{lof}{figure}{%
				\protect\numberline{\thefigure}%
				Representative skill-level cognitive trajectories generated by DIKT.%
			}
			\noindent
			\textbf{Fig.~\thefigure.}
			Representative skill-level cognitive trajectories generated by DIKT.
			The trajectories remain bounded in $[0,1]$ and exhibit smooth but
			heterogeneous evolution patterns across skills.
			\par
		}
	\end{minipage}
	\par
	\vspace{6pt}
	
	As shown in Fig.~\ref{fig:knowledge_trajectory}, the estimated cognitive
	states remain within the intended range of $[0,1]$, which is consistent
	with the design of the cognitive-state decoder. The trajectories also
	evolve smoothly over time rather than exhibiting abrupt fluctuations
	after individual interactions, thereby providing a stable representation
	of the estimated learning process.
	
	Different skills exhibit heterogeneous evolution patterns. Some
	trajectories increase rapidly and gradually approach higher cognitive
	levels, whereas others evolve more slowly or remain at relatively low
	levels. These differences indicate that DIKT can distinguish the
	estimated learning dynamics of individual skills instead of forcing all
	skills to follow a common evolution pattern. Compared with opaque hidden
	representations, the resulting trajectories provide a more direct view
	of the student's estimated strengths, weaknesses, and learning progress
	at the skill level.
	
	In addition to temporal cognitive trajectories, DIKT provides
	exercise-specific attribution scores for the skills required by a target
	exercise. These scores provide an exercise-specific view of the relative
	influence of the required skills on the current prediction.
	
	\par
	\vspace{5pt}
	\noindent
	\begin{minipage}{\columnwidth}
		\centering
		\safeincludegraphics[
		width=\linewidth
		]{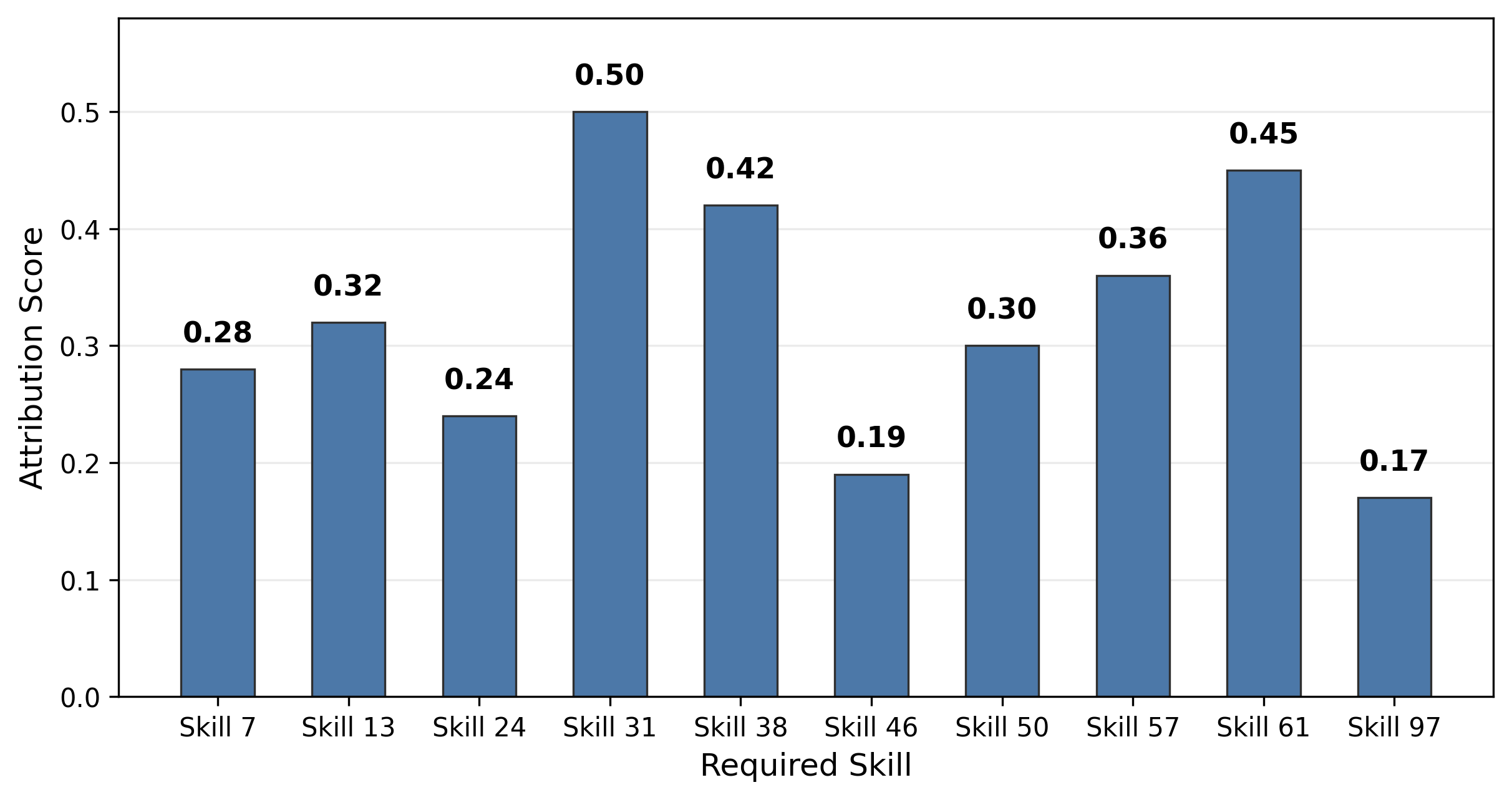}
		\vspace{2pt}
		
		{\normalfont\small
			\setlength{\parindent}{0pt}
			\refstepcounter{figure}
			\label{fig:skill_attribution}
			\addcontentsline{lof}{figure}{%
				\protect\numberline{\thefigure}%
				Exercise-specific attribution scores for the required skills.%
			}
			\noindent
			\textbf{Fig.~\thefigure.}
			Exercise-specific attribution scores for the skills required by a
			representative target exercise. Higher scores indicate stronger
			model-estimated constraints on the predicted response.
			\par
		}
	\end{minipage}
	\par
	\vspace{6pt}
	
	As shown in Fig.~\ref{fig:skill_attribution}, the attribution scores vary
	across the required skills, indicating that the model assigns different
	levels of importance to their contributions to the current prediction.
	Among them, Skill 31 receives the highest attribution score, suggesting
	that it places the strongest model-estimated constraint on the predicted
	success of the target exercise.
	
	These results demonstrate that DIKT can identify the skills that most
	strongly constrain a specific prediction rather than treating all
	associated skills as equally important. Together with the cognitive-state
	trajectories, this attribution analysis provides both temporal and
	exercise-specific diagnostic information.

	\subsection{Diagnostic Analysis (RQ3)}
	\label{subsec:diagnostic_analysis}
	
	This subsection examines whether the learned cognitive states can support
	model-based diagnostic analysis. Specifically, we conduct two complementary
	analyses: intervention analysis and what-if analysis.

	\subsubsection{Intervention Analysis}
	\label{subsubsec:intervention_analysis}
	
	This experiment examines how the predicted response probability changes when the cognitive state of a selected target skill is hypothetically modified. For the target skill $k$, the corresponding entry $c_{t,k}$ is assigned a series of intervention values within $[0,1]$. Each intervention value produces a corresponding predicted response probability, thereby forming an intervention--response curve that illustrates how the predicted response probability varies with the specified cognitive state.

	\begin{figure*}[!t]
		\centering
		
		\safeincludegraphics[width=0.85\textwidth]{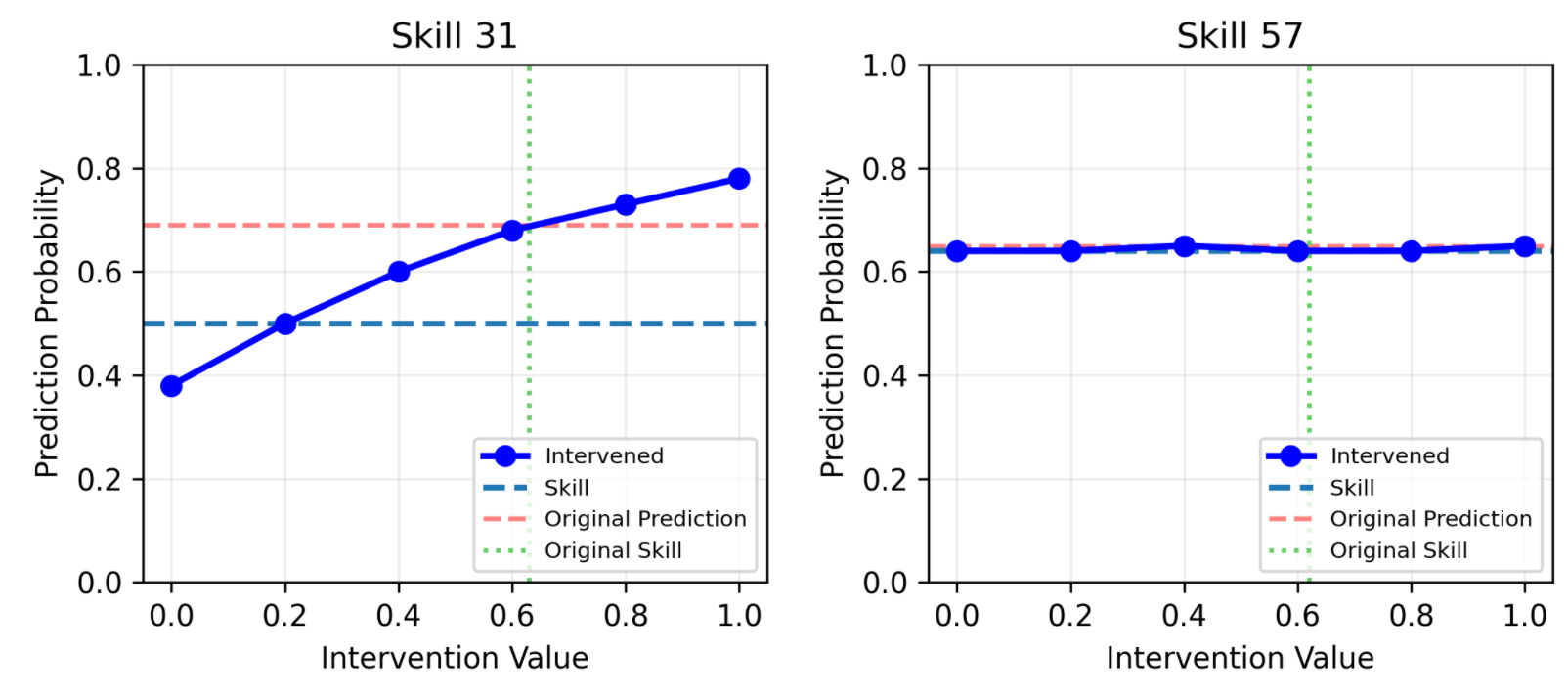}
		
		\vspace{5pt}
		
		\begin{minipage}{\textwidth}
			{\normalfont\small
				
				\refstepcounter{figure}
				\label{fig:intervention_curve}
				\addcontentsline{lof}{figure}{%
					\protect\numberline{\thefigure}Intervention-response curves for representative target skills.%
				}
				
				\noindent
				\textbf{Fig.~\thefigure.}
				Intervention-response curves for representative target skills. The figure shows how the predicted response probability changes as the specified intervention value of a selected skill increases.
				
				\par
			}
		\end{minipage}
		
	\end{figure*}

	As shown in Fig.~\ref{fig:intervention_curve}, different target skills exhibit distinct intervention--response patterns. A non-flat curve indicates that the cognitive state of the selected skill is associated with the final prediction under the learned model. To characterize the model's local sensitivity around the student's current state, we use the intervention efficiency corresponding to the evaluated intervention value closest to the student's current cognitive estimate. For Skill 31, a small increase in the intervention value produces a noticeable improvement in the predicted response probability, with a local intervention efficiency of approximately 0.33, indicating that the prediction is sensitive to this cognitive dimension. In contrast, the nearly flat curve of Skill 57 is associated with a local intervention efficiency close to 0.00, suggesting that this skill has limited immediate influence on the target prediction.

	The intervention analysis reported above can indicate which skill-level cognitive state changes are more likely to improve the predicted performance on a target exercise, thereby providing a quantitative diagnostic means of examining the relationship between skill cognitive and response prediction. In practical applications, downstream diagnostic tools can adopt intervention analysis to design additional indicators for more comprehensive assessment; however, the design and evaluation of such indicators are beyond the scope of this experiment. This type of analysis is generally not available from conventional knowledge tracing models.

	\subsubsection{What-if Analysis}
	\label{subsubsec:whatif_analysis}
	
	In addition to direct cognitive state intervention, DIKT supports what-if analysis. Instead of directly overwriting cognitive values, the what-if module perturbs earlier effective learning-duration dynamics and recomputes the subsequent cognitive trajectory. While the previous intervention analysis focuses on how a hypothetical cognitive change affects the predicted response probability, the what-if analysis further examines how such a perturbation propagates through subsequent skill-level cognitive evolution.
	
	\begin{figure*}[!t]
		\centering
		
		\safeincludegraphics[width=\textwidth]{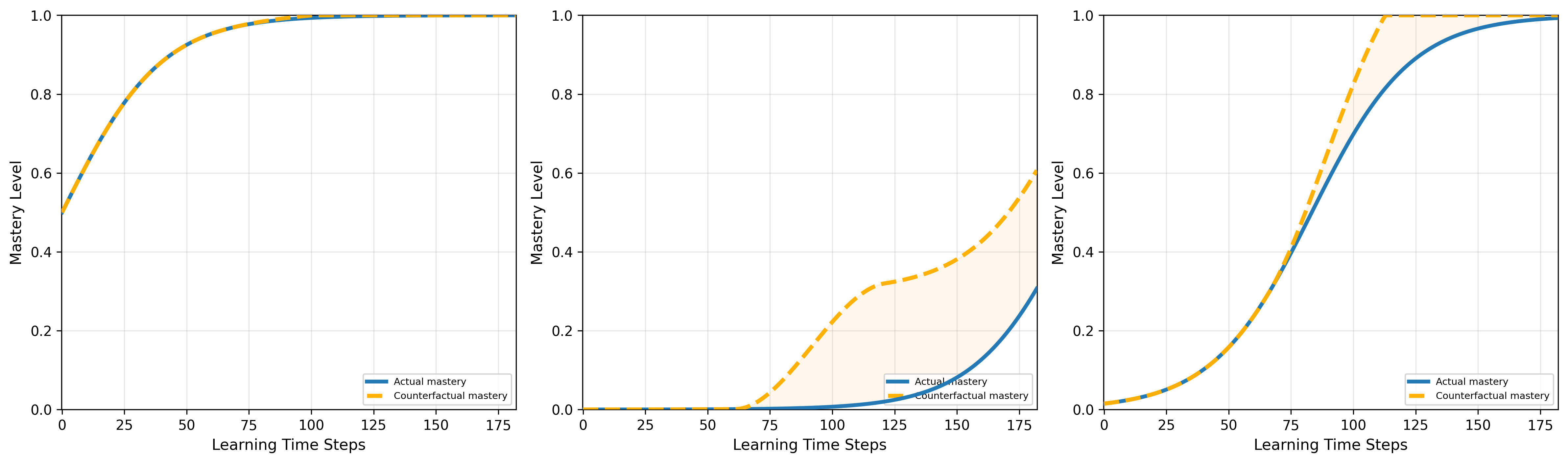}
		
		\vspace{5pt}
		
		\begin{minipage}{\textwidth}
			{\normalfont\small
				
				\refstepcounter{figure}
				\label{fig:whatif_representative_cases}
				\addcontentsline{lof}{figure}{%
					\protect\numberline{\thefigure}Representative what-if cognitive trajectories under different learning states.%
				}
				
				\noindent
				\textbf{Fig.~\thefigure.}
				Representative what-if cognitive trajectories under different learning states. The three cases correspond to a high-cognitive state, a low-cognitive state, and a low-to-high transition state, respectively.
				
				\par
			}
		\end{minipage}
		
	\end{figure*}

	Fig.~\ref{fig:whatif_representative_cases} presents representative what-if cases under different cognitive states. In the high-cognitive case, the original and what-if trajectories almost overlap throughout the sequence, resulting in a $D_{L2}$ value of $0.0020$. This small value suggests that the hypothetical perturbation has limited downstream influence when the target skill is already close to saturation. In the low-cognitive case, the original trajectory remains at a low level, whereas the what-if trajectory gradually diverges after the perturbation and reaches a higher cognitive level over time. The corresponding $D_{L2}$ value is $0.2008$, which is the largest among the three cases. This indicates that skills with low cognitive-state estimates may be more sensitive to changes in effective learning dynamics and that an early positive perturbation can induce a sustained change in subsequent cognitive evolution. In the low-to-high transition case, the what-if trajectory initially deviates from the original trajectory, but the gap gradually narrows as the original trajectory improves over time. The resulting $D_{L2}$ value is $0.0685$, which is larger than that of the high-cognitive case but substantially smaller than that of the low-cognitive case. This suggests that when a skill already exhibits strong subsequent growth, the long-term marginal effect of the perturbation becomes limited.

	The what-if analysis capability has important potential for intelligent tutoring systems. By propagating hypothetical perturbations through the learned cognitive trajectories and performing subsequent response prediction, the model can evaluate the potential effects of different learning scenarios, thereby supporting the selection of more effective personalized learning paths. Such model-based trajectory analysis and predictive decision support are generally beyond the capability of conventional knowledge tracing models.

	\subsection{Ablation Study (RQ4)}
	\label{subsec:ablation_study}
	
	This subsection evaluates the contributions of the main components of
	DIKT to predictive performance and interpretability. Following the
	ablation settings summarized in Table~\ref{tab:ablation_updated}, we
	compare the full DIKT model with seven variants: DIKT-NM removes the
	cognitive-state module; DIKT-DA replaces Bayesian attention with
	deterministic multi-head attention; DIKT-NC removes CVRNN-based
	latent-state modeling; DIKT-NA removes the auxiliary attention-based
	prediction loss $\mathcal{L}_{\mathrm{attn}}$; DIKT-NMo removes the
	monotonicity regularization term $\mathcal{L}_{\mathrm{mono}}$; DIKT-NS
	removes the scale and smoothness regularization term
	$\mathcal{L}_{\mathrm{scale}}$; and DIKT-MLP replaces the multiplicative
	skill-wise prediction layer with a direct MLP prediction head.

	\begin{table*}[!t]
		\centering
		\begin{minipage}{0.92\textwidth}
			{\normalfont\small\raggedright
				
				\refstepcounter{table}
				\label{tab:ablation_updated}
				\addcontentsline{lot}{table}{%
					\protect\numberline{\thetable}Performance comparison of the seven ablation variants and the full DIKT model.%
				}
				
				\noindent\textbf{Table~\thetable}\par
				\vspace{2pt}
				\noindent Performance comparison of the seven ablation variants and the full DIKT model.\par
				\vspace{5pt}
				
				\renewcommand{\arraystretch}{1.08}
				
				\begin{tabular*}{\linewidth}{@{\extracolsep{\fill}}lcccccc@{}}
					\toprule
					\multirow{2}{*}{Methods}
					& \multicolumn{2}{c}{ASSIST09}
					& \multicolumn{2}{c}{ASSIST17}
					& \multicolumn{2}{c}{Algebra2005} \\
					\cmidrule(lr){2-3}
					\cmidrule(lr){4-5}
					\cmidrule(lr){6-7}
					& AUC & ACC & AUC & ACC & AUC & ACC \\
					\midrule
					DIKT-NM  & 0.7644 & 0.7406 & 0.7033 & 0.6841 & 0.7382 & 0.8349 \\
					DIKT-DA  & 0.7169 & 0.7665 & 0.7860 & 0.7279 & 0.7984 & 0.8421 \\
					DIKT-NC  & 0.6860 & 0.7613 & 0.7545 & 0.7178 & 0.7934 & 0.8393 \\
					DIKT-NA  & 0.6291 & 0.7433 & 0.7478 & 0.7142 & 0.7777 & 0.8375 \\
					DIKT-NMo & 0.6781 & 0.7615 & 0.7546 & 0.7185 & 0.7893 & 0.8419 \\
					DIKT-NS  & 0.6568 & 0.7643 & 0.7552 & 0.7169 & 0.7911 & 0.8423 \\
					DIKT-MLP & 0.7487 & 0.7315 & 0.7372 & 0.7082 & 0.7953 & 0.8425 \\
					\midrule
					DIKT
					& \textbf{0.8272}
					& \textbf{0.7761}
					& \textbf{0.7925}
					& \textbf{0.7313}
					& \textbf{0.8485}
					& \textbf{0.8598} \\
					\bottomrule
				\end{tabular*}
				
			}
		\end{minipage}
	\end{table*}

	As shown in Table~\ref{tab:ablation_updated}, all ablated variants show different degrees of performance degradation compared with the full DIKT model. DIKT-NM performs worse than DIKT, indicating that the explicit cognitive state module contributes not only to interpretability but also to response prediction. DIKT-DA also leads to a decrease in performance, suggesting that Bayesian attention is more suitable than deterministic attention for modeling uncertainty in noisy educational interaction sequences. In addition, the performance drop of DIKT-NC shows that CVRNN-based latent-state modeling is important for capturing heterogeneous and dynamic learning processes.
	
	The results of DIKT-NA, DIKT-NMo, and DIKT-NS further demonstrate the contribution of the auxiliary objectives and regularization terms. Removing the auxiliary attention loss $\mathcal{L}_{\mathrm{attn}}$ weakens the supervision of attention-based prediction, while removing the monotonicity regularization $\mathcal{L}_{\mathrm{mono}}$ or the scale regularization $\mathcal{L}_{\mathrm{scale}}$ reduces the educational plausibility and stability of the learned cognitive trajectories. Moreover, DIKT-MLP performs worse than the full DIKT model, which supports the value of the multiplicative skill-wise prediction layer in preserving an explicit pathway from skill-level cognitive state to response prediction.

	\section{Conclusions and future work}
	\label{sec:conclusion}
	
	This paper proposed DIKT, a Diagnostic and Interpretable Knowledge
	Tracing framework that integrates response prediction, skill-level
	diagnosis, and model-based learning analysis. By combining Bayesian
	multi-head attention, CVRNN-based latent-state inference, a
	logistic-growth cognitive-state decoder, and a multiplicative
	skill-wise prediction mechanism, DIKT explicitly models bounded and
	smooth skill-level cognitive trajectories. The skill-wise prediction
	mechanism further provides skill-wise success probabilities and
	attribution scores, thereby establishing an explicit pathway from
	skill-level cognitive-state estimates to the final item-level response
	prediction. Based on the learned cognitive states, DIKT also supports
	direct cognitive-state intervention and effective-learning-duration-based
	what-if analysis for examining the immediate and temporal effects of
	hypothetical learning changes on model predictions.
	
	Experiments on three benchmark datasets demonstrate that DIKT achieves
	strong predictive performance while producing distinguishable
	skill-level cognitive trajectories and exercise-specific attribution
	scores. The intervention results show that the predicted response
	exhibits different sensitivities to changes in different skill-level
	cognitive states, whereas the what-if results illustrate how changes in
	effective learning duration may propagate through subsequent cognitive
	trajectories. These results demonstrate that DIKT can provide diagnostic
	information from both the current prediction and temporal evolution
	perspectives, thereby extending knowledge tracing beyond response
	prediction. In future work, we will incorporate richer educational and
	contextual information, improve the visualization and practical usability
	of skill-level cognitive trajectories and attribution results, and
	further evaluate        the diagnostic value of DIKT through expert assessment
	and real instructional applications.
	\printcredits
	\section*{Declaration of Competing Interest}
	The authors declare that they have no known competing financial interests or personal relationships that could have appeared to influence the work reported in this paper.
	
	\section*{Data availability}
	Data will be made available on request.
	
	\section*{Acknowledgements}
	This work is supported by the National Natural Science Foundation of China (Grants No. 12035006 and
	No. 12205310). Computations are performed at the
	High-Performance Computing Center of Wuhan University of Science and Technology.

\end{document}